\documentclass[letterpaper]{article} 
\usepackage[]{aaai23}
\usepackage{times}  
\usepackage{helvet}  
\usepackage{courier}  
\usepackage[hyphens]{url}  
\usepackage{graphicx} 
\usepackage{natbib}  
\usepackage{caption} 
\usepackage{algorithm}

\usepackage{newfloat}
\usepackage{listings}
\DeclareCaptionStyle{ruled}{labelfont=normalfont,labelsep=colon,strut=off} 
\floatstyle{ruled}
\newfloat{listing}{tb}{lst}{}
\floatname{listing}{Listing}
\usepackage{xcolor}         
\usepackage{xspace}

\usepackage{amssymb}
\usepackage{amsmath}
\usepackage{algpseudocode}
\usepackage{subfigmat}
\usepackage{paralist, tabularx}
\usepackage{array}
\usepackage{multirow}

\newcommand{\clientbox}[1]{\colorbox{myblue!30}{#1}}
\newcommand{\serverbox}[1]{\colorbox{mygreen!30}{#1}}

\newcommand{\mtasks}[1]{\colorbox{myred!30}{#1}}

\definecolor{mygreen}{RGB}{77,175,74}
\definecolor{myblue}{RGB}{55,126,184}
\definecolor{skyblue}{RGB}{117,187,253}
\definecolor{myred}{RGB}{228,26,28}

\newcommand{\paragraphb}[1]{\noindent{\bf #1} }

\newcommand{\XYZ}[0]{\name}
\DeclareMathOperator*{\argmax}{argmax}
\DeclareMathOperator*{\argmin}{argmin}
\usepackage{tikz}
\newcommand*\circled[1]{\tikz[baseline=(char.base)]{
            \node[shape=circle,draw,inner sep=.3pt] (char) {#1};}}

\newcommand{\name}{{E2FL}\xspace}

\title{\name: Equal and Equitable Federated Learning}
\author{
    Hamid Mozaffari, Amir Houmansadr
}
\affiliations{
    University of Massachusetts Amherst\\
    hamid@cs.umass.edu, amir@cs.umass.edu

}

\usepackage{bibentry}

\begin{document}

\maketitle

\begin{abstract}
  Federated Learning (FL) enables data owners train a shared global model without sharing their private data. 
  Unfortunately, FL is susceptible to an intrinsic fairness issue: due to the heterogeneity of clients' data distributions, the  final FL model  can give disproportionate advantages across the participating clients. 
  In this work, we present Equal and Equitable Federated Learning (\name) to produce fair federated learning models by preserving two main fairness properties,   equity and equality, \emph{concurrently}.
 We validate the efficiency and fairness of \name in different real-world FL applications, and show that \name outperforms existing baselines in terms of the resulting efficiency, fairness of different groups, and fairness among all individual clients. 
\end{abstract}

\section{Introduction}
Federated Learning (FL) is an emerging AI technology where \emph{clients} collaborate to train a shared model, called the \emph{global model}, without explicitly sharing their local training data. FL training involves a \emph{server} which collects model updates from selected FL clients in each round of training, and uses them to update the global model. 
In FL, the performance of the global model varies across the clients due to heterogeneity in the data that each client owns. 
This concern is called \emph{representation disparity}~\cite{hashimoto2018fairness} and results in unfair performance gaps for the participating clients. That is, although the accuracy may be high on average, some tail user whose data distribution differs from the majority of the clients is likely to  receive a much lower performance compared to the average. 

In this work, we look at FL fairness with two different lenses: \textbf{a)} \emph{Equality}: whose goal is providing similar performances for all  individual clients; \textbf{b)} \emph{Equity}: whose goal is providing similar performances across all  groups of clients (i.e., groups of majority and minority), where 
a group is defined as a set of clients with similar data distributions. 
The key question we try to answer is: \emph{Can we design an efficient federated learning algorithm that achieves both equality and equity concurrently?}


Due to the heterogeneity in clients’ data distributions, one single model cannot represent all the clients equally. 
There is a \emph{trade-off} between training one global model and multiple global models; if we  train one global model all the clients can utilize each other's knowledge, however it will be biased towards whom that have the majority of the population. On the other hand, if we train multiple  models (e.g., as in IFCA~\cite{ghosh2020efficient}, HypCluster~\cite{mansour2020three} and MOCHA~\cite{Smith17}), we improve fairness, but each global model will lose the knowledge from excluded clients.  
To get the best of both worlds, we present \textbf{E}qual and \textbf{E}quitable \textbf{F}ederated \textbf{L}earning (\name), a novel FL algorithm to achieve both equality and equity.
In \name, we train multiple global models, but in each round we combine all of the models into one global model to take advantage of the knowledge of all client groups. 


The key insight used in \name is converting the problem of model weight optimization (in standard FL) to the problem of ranking model edges (a technique recently proposed in \cite{mozaffari2021frl}). 
Therefore, in each round of \name training, the clients and the server exchange rankings for the edges of a randomly initialized neural network (called \emph{supernetwork}), as opposed to exchanging parameter gradients.
More specifically, each \name client computes the importance of the edges within a randomly initialized neural network (called supernetowrk) on their local data, represented by a ranking vector. Next, \name server uses a majority voting mechanism to aggregate the collected local rankings into multiple global rankings based on the index of group they belong to. Finally, the \name server aggregates all the group rankings into one global ranking for next round of training. Applying the majority vote on the group rankings instead of all the local rankings helps  \name  enforce equity because each group has an equal vote to influence the global model. To provide equality in \name, if a client wants to use the model in a downstream task, they use their own group global ranking, instead of the global ranking, which is a better representation model for the client and its groupmates. 

Our ranking-based FL training  enables attractive fairness properties, as shown through our experiments, which is intuitively due to the following reason: 
In rank-based federated learning, each client computes a local ranking (i.e., a permutation of integers $\in [1,d]$ where $d$ is the layer size), so each local ranking has a fixed norm (i.e., $\sqrt{1^2+2^2+...+d^2}$). 
This fixed norm of local updates makes the rank aggregation more fair as each local ranking has the same impact on the aggregated global ranking.
On the other hand, in standard FL, when the server aggregates the local model updates into the global model, each local update has a different impact on the global model (because of their different $l_2$ norms). For example in FedAvg, the server averages the parameter updates for the $d$ dimensions, therefore a large parameter update has more influence on the final average compared to a small parameter update.

\paragraphb{\name when the group IDs are unknown.} In many applications, clients may be unaware of their protected attributes (i.e., the group they belong to). We propose one approach on server-side and three approaches on client-side for inferring  group IDs. To  infer the group IDs on the server-side, we propose to use a rank clustering approach to cluster  clients into groups.
Moreover, a client can also infer its group IDs by picking the right group based on their local training data. Using rankings allows us to exchange only the binary masks produced by each group ranking which lowers the communication cost compared to prior works.
Each client can pick the right binary mask based on three approaches. First, each client can pick the binary mask that produces the smallest loss. Binary masks also enable the clients to find their matching group by a new novel idea from~\cite{wortsman2020supermasks}, where clients can infer the group ID using gradient based optimization to find a linear superposition of learned masks which minimizes the output entropy. We propose two variants of this approach, one based on a binary search and the other using OneShot optimization.

\paragraphb{\em Empirical results:}
We experiment with three datasets in real-world heterogeneous FL settings and show that \name can help clients from both majority and minority groups, while q-FFL~\cite{li2019fair}, state-of-the-art fair FL, improves equality by helping the majorities and ignoring the minorities.
Since there are no existing distributed datasets containing different groups of clients with different data distributions, 
we create a new dataset, called FairMNISTRotate, to evaluate  equality and equity in FL applications. FairMNISTRotate represents 10 different handwriting styles, produced by rotating samples from the MNIST dataset. 
For FairMNISTRotate,
 q-FFL reduces the variance of accuracies for all  clients by 4\% while it increases the variance between groups by 81\% compared to FedAvg. On the other hand, our algorithm  reduces the variance of both clients and groups by 93\% and 95\% respectively compared to FedAvg. That is, \name improves both equity and equality, unlike prior fair FL algorithms. 


\section{Background}

\paragraph{Federated Learning:}
In FL~\citep{mcmahan2017communication, kairouz2019advances, konevcny2016federated}, 
$N$ clients collaborate to train a global model  without directly sharing their data. 
In  round $t$, the service provider (server) selects $n$ out of $N$ total clients and  sends them the most recent global model $\theta^t$. Each client trains a local model for $E$ local epochs on their data starting from the $\theta^t$ using stochastic gradient descent (SGD). Then the client  sends back the calculated gradients to the server. The server then aggregates the collected gradients and  updates the global model for the next round.
%


\paragraph{Rank-based Federated  Learning:}\label{sec:Rank-based}
Federated Rank Learning (FRL)~\cite{mozaffari2021frl} is an approach to perform FL that is built on a novel learning paradigm called \emph{supermasks}~\citep{DBLP:conf/nips/ZhouLLY19, ramanujan2020what}.
Specifically, in FRL, clients collaborate  to find a \emph{subnetwork} within a \emph{randomly initialized} neural network (the supernetwork), where this is in contrast to conventional FL where clients collaborate to \emph{train} neural network parameter weights.
The goal of training in FRL is to collaboratively identify a supermask $M_g$, which is a binary mask of 1's and 0's, that is superimposed on the random neural network to obtain the final subnetwork, i.e., $\theta^w \bigodot M_g$ where $\theta^w$ is showing the weight parameters for supernetwork.
$M_g$ contains the edges of top $k$ ranks, i.e., edges in top $k$ ranks (layer-wise) get '1' in the binary mask, and others get '0' in the mask. We use $k=50\%$ in this work to find a subnetwork of 50\% of the original size.
The subnetwork is then used for downstream tasks, e.g., image classification, hence it is equivalent to the global model in conventional FL. Note that in entire FRL training, weights of the network ($\theta^w$) do not change.

More specifically, 
each  FRL client computes the importance of the edges of the supernetwork based on their local data. The importance of the edges is represented as a ranking vector. 
Each FRL client will use the edge-popup algorithm~\citep{ramanujan2020what} and their data to compute their local rankings (the edge-popup algorithm aims at learning which edges in the random network are more important over the other edges by minimizing the loss of the subnetwork on their local data). 
Each client then will send their local edge ranking to the server.
Finally, the FRL server uses a \emph{voting mechanism} to aggregate client rankings into a  supermask, which represents which edges of the random neural network (the supernetwork) will form the global subnetwork. 
Specifically, the FRL server uses Borda count rank aggregation method~\cite{emerson2013original} where it gives a reputation to each edge for each ranking, sums the reputations, and sorts them from least to most to find the global ranking.
We defer further details of FRL and edge-popup Algorithms to Appendix~\ref{sec:FRL}.


\section{Fairness Using Two Lenses: Equity and Equality}
Fairness in FL can be evaluated from two main perspectives:
a) \emph{Equality} which is fairness between individuals and b) \emph{equity} which is fairness between groups. A group is a set of individuals with the same protected attribute. 
The protected attribute may be known to the clients,  e.g., race, gender, or age. Alternatively, clients may be unaware of their particular group,  e.g., handwriting style (as this needs  clustering clients into groups by someone who has samples from all clients).

\begin{figure}[htb!]
    \centering
    \includegraphics[width=0.49\textwidth]{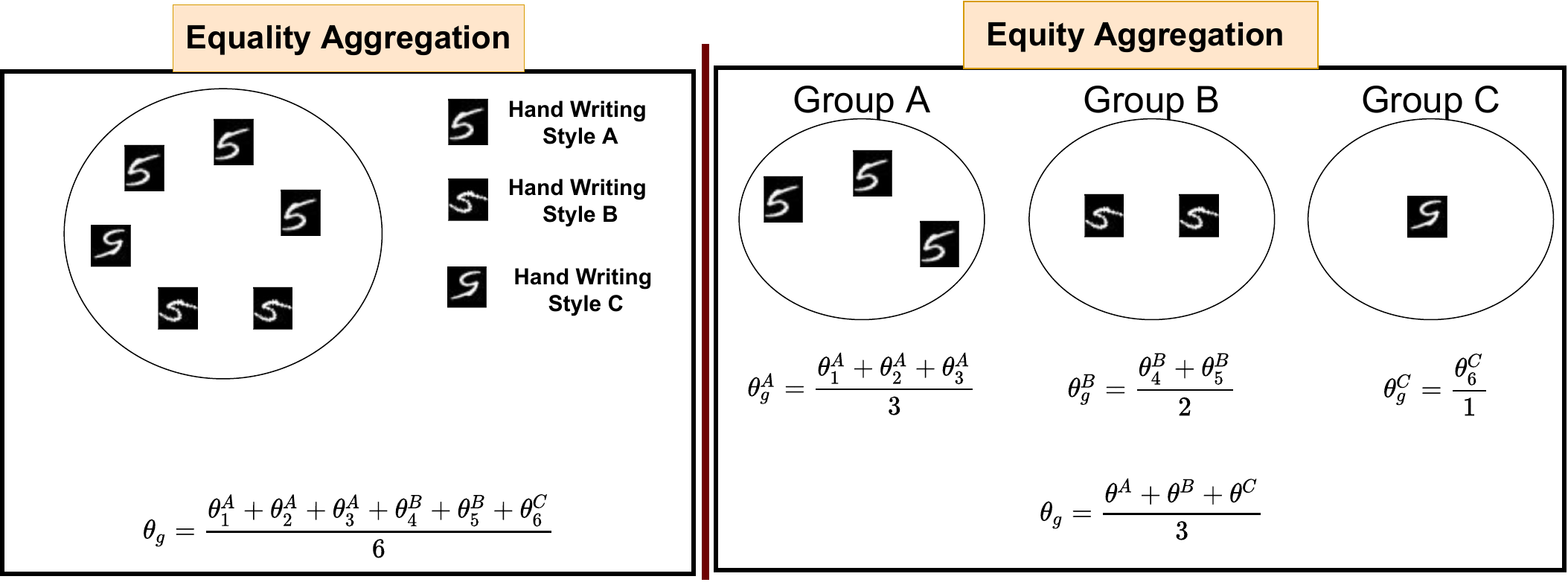} 
    \vspace{-15pt}
    \caption{An example showing two different FL systems with two goals: equality (on left) and equity (on right).}
    \label{fig:Fairness1}
    \vspace{-10pt}
\end{figure}

Figure~\ref{fig:Fairness1} shows an example of two FL systems where six clients want to learn a global model for prediction of handwritten digits.
These clients have three handwriting styles: \textbf{(A)} normal handwriting style, \textbf{(B)} a little bit rotated handwriting, and \textbf{(C)} 180 degree rotated handwriting (upside-down).
We consider each model update ($\theta^q_u$ for client $u$ in group $q$) has the same effect on updating the global model, so each client update is like a vote. 
In this example, group A has the majority of the voters, and group B and C are in minorities. 
The left part of figure shows an FL in which the goal is providing equality, so we give same chance (one vote) to each client to change the final model by an aggregation such as averaging (e.g., what we have in FedAvg). In this setting, the majority group with higher population (group A) has more influence on the final vote. 
On the other hand, the right part shows an FL in which the goal is providing equity. In this setting, first we aggregate the votes inside each group to find the group votes ($\theta^{A}_g, \theta^{B}_g, \theta^{C}_g$), and then aggregate the group votes to produce the final model. 
In this setting, each client has the same chance (one vote) to influence its own group vote, and finally each group of voters have the same chance (one vote) to influence the final vote.
We define two aspects of fairness in FL as follows:

\paragraphb{Definition 1 (Equality: User-level Fairness):} 
Trained global model $\theta$ is more \emph{equalized} when its performance is more uniform across the individual clients participating in FL, i.e., when  STD$\{F_u(\theta)\}_{u \in [N]}$ is smaller where STD$\{.\}$ is the standard deviation, and $F_u(.)$ denotes the local objective function of client $u$ from $N$ clients. 
Existing fair federated learning literature~\cite{li2019fair, li2020ditto, Smith17, hashimoto2018fairness, zhang2020personalized, mohri2019agnostic, yu2020salvaging} use this definition in their designs.    

\paragraphb{Definition 2 (Equity: Group-level Fairness):}
Trained global model $\theta$ is more \emph{equitable} when its performance is more uniform across the groups, i.e., when 
STD$\{$Avg$\{F_u(\theta)\}_{u \in [q]}\}_{q \in [Q]}$ is smaller where AVG$\{\}_{u \in [q]}$ denotes the average of performances for all the individual clients in the $q$th group, and there are $Q$ total groups. 


In \name, our goal is providing both equity and equality. To provide equity, an individual client has one vote in their group, and each group has one vote among all the groups. To provide equality, we allow the clients in each group use their group model which represents their training data better.

\vspace{-3pt}
\section{\XYZ{}: Design}
In this section, we provide the design of our Equal and Equitable Federated Learning (\name) algorithm. We first describe how \name
provides equity and then discuss how it provides equality. 
The intuition behind \XYZ{} is to train multiple global models, and first perform a majority vote among the clients' model updates in each group, and then another majority vote among the group models to find the global model.
Algorithm~\ref{alg:FFRL} describes  \XYZ{} training.



\begin{algorithm}[h]
\footnotesize
\caption{Equal and Equitable Federated Learning (\name) Algorithm.}\label{alg:FFRL}
\begin{algorithmic}[1]
\footnotesize
\State \textbf{Input:} number of FL rounds $T$, number of local epochs $E$, number of selected users in each round $n$, number of groups $Q$, seed $\textsc{seed}$, learning rate $\eta$, subnetwork size $k$\%
\State $\theta^s, \theta^w \gets$ Initialize random scores and weights of global model $\theta$ using \textsc{seed}
\State $R_{g}^{1} \gets \textsc{ArgSort}(\theta^s)$ \Comment{Sort the initial scores and obtain initial global rankings}
\For{$t \in [1,T]$}
    \State $U \gets$ set of $n$ randomly selected clients out of $N$ total clients
    \For{$u$ in $U$}
        \State $\theta^s, \theta^w \gets $ Initialize scores and weights using \textsc{seed}
        \State If $q$ (group ID) is not known, use Algorithms~\ref{alg:IDEKmeans} and~\ref{alg:IDE} for ID inference (Section~\ref{Sec:main_unkown})
        \State $\theta^s[R_{g}^{t}] \gets \textsc{sort}(\theta^s)$ \Comment{Reorder the scores based on the global ranking}
        \State $\theta_{u}^s \gets$ Edge-PopUp($E, D_u^{tr}, \theta^w, \theta^s, k, \eta$) \Comment{Train local scores on the local training data}
        \State $R_{u, q}^{t} \gets \textsc{ArgSort}(\theta_{u}^s)$ \Comment{Ranking of the client $u$ with estimated group ID: $q$}
        \State \textbf{return} $R_{u,q}^{t}$
    \EndFor
    \State \serverbox{$R_{g, q\in[Q]}^{t+1} \gets \textsc{Vote}(R_{u \in U, q\in[Q]}^{t})$ } \Comment{Majority vote aggregation inside each group}
     \State \serverbox{$R_{g}^{t+1} \gets \textsc{Vote}(R_{g, q\in[Q]}^{t+1})$} \Comment{Majority vote aggregation among all the groups}
\EndFor
\Function{\serverbox{Vote}}{$R_{\{u \in U\}}$ }:
    \State $V \gets \textsc{ArgSort}(R_{\{u \in U\}})$ \Comment{Reputation of each edge in each local ranking}
    \State $A \gets \textsc{Sum}(V)$ \Comment{Sum the reputations}
    \State \textbf{return} $\textsc{ArgSort}(A)$ \Comment{Order of the reputations}
\EndFunction
\end{algorithmic}
\end{algorithm}

The key insight used in \name is converting the problem of model weight optimization (in standard FL) to the problem of ranking model edges~ \cite{mozaffari2021frl}. 
In Section~\ref{sec:Rank-based}, we explained how an FL works by training on parameter ranks. 
In \name, each local update (one vote) is a ranking, i.e., a permutation of integers $\in [1, d]$ where $d$ is the size of network layer.
We use rankings because of their intrinsic fairness feature: In rank aggregation, each local ranking has the same impact on the aggregated global ranking. In rank-based FL, all the local rankings are bounded to be a permutation of unique integers $\in [1, d]$. 
For example, for a network layer with $d=3$ parameters, there are only $3!$ possible permutations for local ranking (${[1,2,3], \dots, [3,2,1]}$).
However, in existing standard FL designs, the local updates ($\in \mathbb{R}^d$) have different impacts on the aggregated global model because the direction and magnitude of each parameter update is not bounded to other parameters. 


In \name, different FL clients gather together to learn a global model, but each one of them belongs to a different group (which could be considered known or unknown). In this section, we assume the clients know their group IDs, and in Section~\ref{Sec:main_unkown}, we explain how the clients can infer their group IDs using the features of rankings when the groups are unknown. In \XYZ{}, the server trains multiple global rankings, each one belonging to a different group. These global group rankings are showing different orders of importance of same supernetwork for different groups from least to most important edges. Each client participates in the training of their group model by sending the local ranking they have. For aggregation, the server performs a majority vote among the local rankings (local votes) in each group, and then performs another majority vote among global group rankings (group votes) to find the global model for the next round (i.e., global ranking that clients will start their training for next \name round).

\paragraphb{Edge-PopUp Algorithm}
The edge-popup (EP) algorithm~\citep{ramanujan2020what} is an optimization method to find supermasks within a large, randomly initialized neural network, i.e., a supernetwork, with performances close to the fully trained supernetwork.
 EP algorithm does not train the weights of the network, instead only decides  the set of edges to keep and removes the rest of the edges (i.e., pop).
Specifically, EP algorithm assigns a positive score  ($\theta^s$) to each of the edges in the supernetwork and updates it.
In \name, each client learns its local scores $\theta_{u}^s$ by using EP on its local data for $E$ local epochs starting from the global scores $\theta^s $.
On forward pass, it selects the top $k\%$ edges with highest scores, where $k$ is the percentage of the total number of edges in the supernetwork that will remain in the final subnetwork. On the backward pass, it updates the scores with the straight-through gradient estimator~\citep{bengio2013estimating}. Algorithm~\ref{alg:edgepop} in Appendix~\ref{sec:EP} presents EP algorithm.
 \begin{figure}[h]
    \centering
    \includegraphics[width=0.5\textwidth]{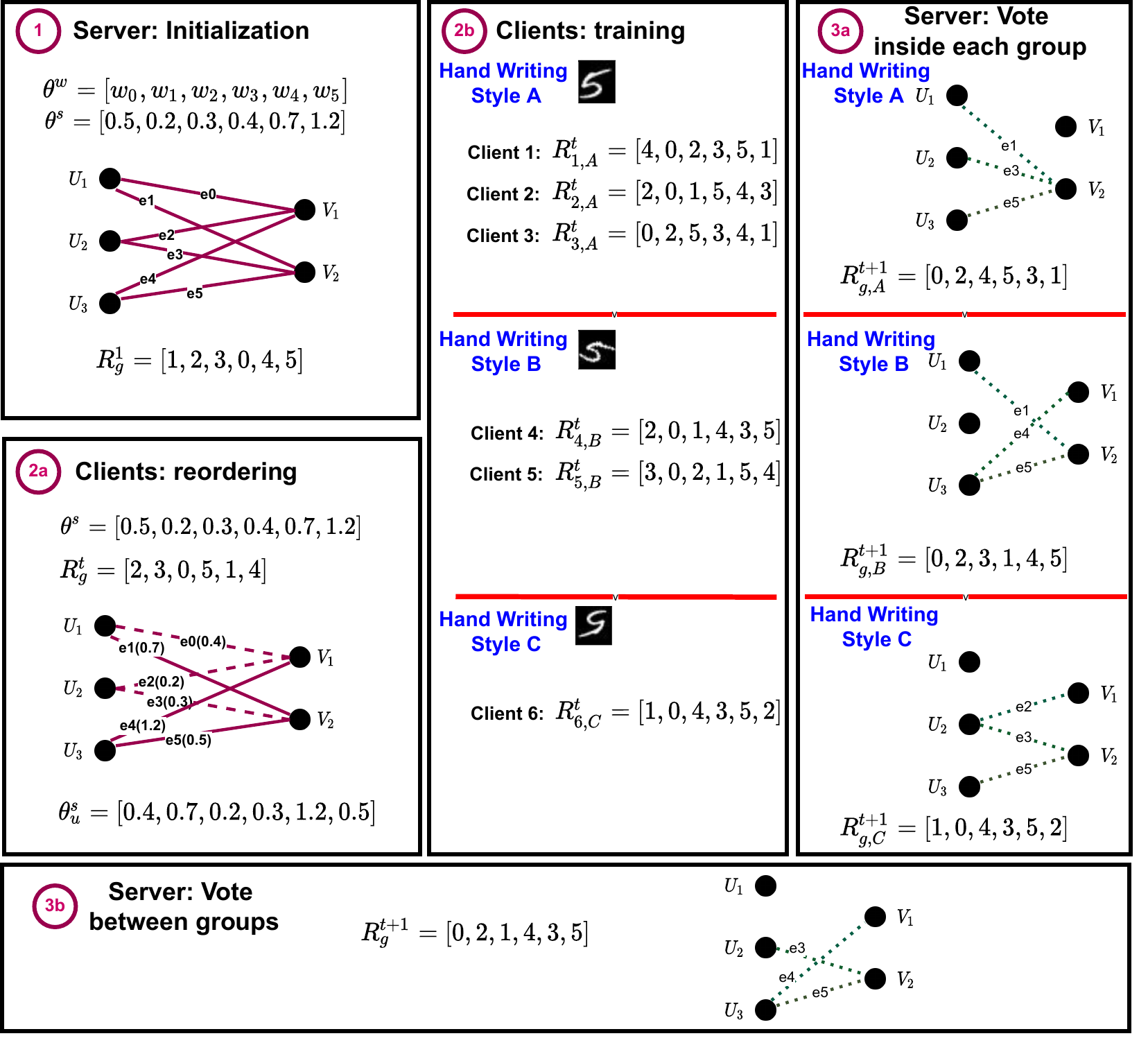}
    \vspace{-20pt}
    \caption{A single \XYZ{} round with six clients from three groups and a network of 6 edges. Note that all the operations in \name training are performed in a layer-wise manner. 
    }
    \label{fig:fflr_design}
    \vspace{-5pt}
\end{figure}

We detail a round of \XYZ{} training and depict it in Figure~\ref{fig:fflr_design}, where we use a supernetwork with six edges $e_{i\in[0,5]}$ to demonstrate a single \XYZ{} round and consider six clients $C_{j\in[1,6]}$ from three groups (handwriting style A, B, C) who aim to find a subnetwork of size $k$=50\% of the original supernetwork.


\paragraphb{Server: Initialization Phase (Only for round $t=1$)}:
In the first round, the \XYZ{} server chooses a random seed \textsc{Seed} to generate initial random weights $\theta^w$  and scores $\theta^s$ for the global supernetwork $\theta$; note that, $\theta^w$, $\theta^s$, and \textsc{Seed} remain constant during the entire \XYZ{} training. Next, the \XYZ{} server shares  \textsc{Seed} with  \XYZ{} clients, who can then locally reconstruct the initial weights $\theta^w$ and scores $\theta^s$ using \textsc{Seed}. Figure~\ref{fig:fflr_design}-\circled{1} depicts this step. 
Recall that, the goal of \XYZ{} training is to find the most important edges in $\theta^w$ without changing the weights. 
At the beginning, the \XYZ{} server finds the global rankings of the initial random scores , i.e., $R_g^1= \textsc{ArgSort($\theta^s)$}$.
We define \emph{rankings of a vector} as the indices of elements of vector when the vector is sorted from low to high, which is computed using \textsc{ArgSort} function.

\paragraphb{Clients: Calculating the ranks (For each round $t$)}:
In the $t^{th}$ round, \XYZ{} server shares the global rankings $R_{g}^{t}$ with the clients. 
Each of the clients locally reconstructs the weights $\theta^w$'s and scores $\theta^s$'s using \textsc{seed}.
Then, each \XYZ{} client reorders the random scores based on the global rankings, $R_{g}^{t}$. We depict this in Figure~\ref{fig:fflr_design}-\circled{2a}. For instance, the initial global rankings for this round are $R^t_g=[2,3,0,5,1,4]$, meaning that edge $e_4$ should get the highest score ($s_4=1.2$), and edge $e_2$ should get the lowest score ($s_2=0.2$).

Next, each of the clients uses reordered $\theta_u^s$ and finds a subnetwork within $\theta^w$ using edge-popup algorithm~\citep{ramanujan2020what}; to find a subnetwork, they use their local data and $E$ local epochs. Note that, each iteration of edge-popup algorithm updates the scores $\theta_u^s$.
Then client $u$ computes their local rankings $R_{u}^t$ using the final updated scores and \textsc{Argsort}(.), and sends $R_{u,q}^t$ to the server where $q$ is the group identifier. We will explain the group inference methods we propose in Section~\ref{Sec:main_unkown}. 
Figure~\ref{fig:fflr_design}-\circled{2b} shows, for each client, the local rankings they obtained after finding their local subnetwork. For example, rankings of client $C_1$  are $R^t_{1,A}=[4,0,2,3,5,1]$, i.e., $e_4$ is the least important and $e_1$ is the most important edge for $C_1$. Considering desired subnetwork size to be 50\%, $C_1$ uses edges \{3,5,1\} in their final subnetwork in this round.

\paragraphb{Server: Majority Vote  (For each round $t$)}:
The server receives all the local rankings of the clients, i.e., $\{R^t_{1,A}, R^t_{2,A}, R^t_{3,A}, R^t_{4,B}, R^t_{5,B}, R^t_{6,C}\}$. 
Then, it performs a majority vote over all the local rankings inside each group, i.e., $\{A, B, C\}$. We depict this in Figure~\ref{fig:fflr_design}-\circled{3a}.
Note that, for group $q$, the index $i$ in ${R^{t+1}_{g,q}}$  represents the  importance of the edge $i$th for clients in group $q$.
For instance, in Figure~\ref{fig:fflr_design}-\circled{3a}, rankings of $A$ are $R^t_{g,A}=[0,2,4,5,3,1]$ and rankings of $B$ are $R^t_{g,B}=[0,2,3,1,4,5]$, hence the edge $e_1$ is the most important edge for group $A$, while the edge $e_5$ is the most important edge for group $B$. 
Next, the server performs a majority vote over all the group rankings of different groups $\{R^{t+1}_{g,A}, R^{t+1}_{g,B}, R^{t+1}_{g,C}\}$ to find the global ranking $R_{g}^{t+1}$. We depict this in Figure~\ref{fig:fflr_design}-\circled{3b}.



\paragraphb{\name provides both equity and equality.} Equity is not the main goal of existing distributed learning systems because it can hurt the motivation of the majority of clients to  participate in the FL. If we have a learning algorithm that provides equity, it has this constraint to allow the same contribution from all groups (i.e., majorities and minorities). This comes with the price of reducing the performance of the majorities, which can demotivate them to participate to learn a model. 

\name provides both equity and equality. 
In this algorithm, at the final round of the learning, instead of using the global ranking, each group uses its own group global rankings. The global rankings can provide better performances to the majority groups as they have access to more training data, so they can train better group global rankings. 
For example, a client of handwriting style A will use $f(x, \theta^w \bigodot M^{t}_{g,A})$ in their downstream classification task, where $M^{t}_{g,A}$ is the learned binary mask for group A at FL round $t$, and $\theta^w$ is the random weights (initialized randomly and kept fixed), and $x$ is the test input. Note that in \name and its variants, $M^{t}_{g,q}$ is the supermask trained for group $q$ where for top $k$\% of the top rankings of group ranking $R_{g,q}^t$, we put 1's and we set other masks to 0's. 
\section{\name when Group IDs are Unknown}\label{Sec:main_unkown}
In the previous section, we assumed that the clients know their group IDs.
In this section, we explain the approaches the server or a client can utilize to estimate the group IDs when the groups are unknown.
In this setting, there are federated clients that have small amount of data with no known protected attribute. For instance, people with their own style of handwriting want to learn a global model by learning a local model on the images of their handwriting. The clients have no knowledge about their style as it is not something to be identified. It is not even possible to announce that clients with similar style collaborate with each other. This type of scenario usually happens in cross-device setting, where each user has a small dataset and there are so many clients in the system. 
For the server-side approach, the server clusters the local rankings into different groups and assigns a group ID to each client. For client-side approaches, each client should estimate its own group ID using the binary masks learned so far (Algorithm~\ref{alg:FFRL} line 8).





\paragraphb{Communication Cost of \XYZ{} when group IDs are unknown:} Please note that for finding the best group ID at the client-side, there is no need to send all the group rankings (e.g., $\{R^{t+1}_{g,A}, R^{t+1}_{g,B}, R^{t+1}_{g,C}\}$ in our example) to the clients. As we mentioned before, in \name each ranking (local or global) can be converted to a binary mask of '0's and '1's that is superimposed on the random weights (i.e., supermask). Thus, the server only broadcasts the binary masks of groups (e.g., $\{M^{t+1}_{g,A}, M^{t+1}_{g,B}, M^{t+1}_{g,C}\}$ in our example) to the clients so they can estimate their group ID where they belong to.



\subsection{Server-side:  Rank Clustering}
Working with rankings enables us to design an efficient algorithm to cluster the local rankings. 
Clustering rankings is more efficient than clustering the model weight updates in standard FL. The main reason is that rankings are from a \emph{discrete} space ($\in perm([1,d])$, all the possible permutation of integers $\in [1,d]$ where $d$ is the layer size) while model updates in standard FL are from a \emph{continuous} space ($\in \mathbb{R}^d$).
In this approach, all the clients should learn a local rankings on their local data at the beginning of the learning, and send it to the server. Then the server clusters the rankings into $Q$ clusters to find the group ID of each client. This is just one time clustering, and throughout the \name learning, the server selects the local rankings for different group rank aggregation (majority voting) based on their group IDs (estimated with this approach).

Algorithm~\ref{alg:IDEKmeans} in Appendix~\ref{Sec:unkownalgs} shows how the server can cluster the local rankings of $N$ clients into $Q$ groups. We adapt K-means clustering to cluster the rankings. In this algorithm, at first step (Algorithm~\ref{alg:IDEKmeans} line 3), we are choosing $Q$ random rankings as our initial $Q$ clusters, called centroids. Then we assign the cluster ID of the closest centroid for all the $N$ local rankings (Algorithm~\ref{alg:IDEKmeans} line 7-10). To determine the distance of two rankings, we use Spearman rank distance in which the distance between two rankings of $R_1$ and $R_2$ is $D(R_1, R_2) = \sum_{\ell\in[L]} \sum_{i \in [n_{\ell}]} |R_1[i]-R_2[i]|$ where $R_1[i]$ shows the rank of parameter $i$th in ranking $R_1$, $L$ is the number of layers in the network, and $n_{\ell}$ shows the number of parameters in layer $\ell$.
In next step (Algorithm~\ref{alg:IDEKmeans} line 11), we are updating the centroid of each cluster by applying majority vote on the ranking inside each cluster. We repeat this process for $T$ iterations to find final $Q$ groups of rankings. 



\subsection{Client-side:  Lowest Loss}
In this approach, each client estimates its group ID by choosing the group that its binary mask produces the lowest loss. Thus, in each \name round, the server broadcasts all the binary masks related to existing groups ($M_{g, q \in [Q]}^t$), and each selected client calculates the loss for each binary mask on its training data. Algorithm~\ref{alg:IDE} line 2-4 in Appendix~\ref{Sec:unkownalgs} shows this approach. This approach was used by IFCA~\citep{ghosh2020efficient}, where in their algorithm in each training round, the server broadcasts all the model parameters to clients, and then they can find the lowest loss group. Note that our \name, compared to IFCA, needs $\times 32 (\times 64)$ less download bandwidth in each round because it is working on the binary masks.



\subsection{Client-side: Entropy of the output}
Binary masks enable us to utilize other approaches for group inference. In these approaches~\cite{wortsman2020supermasks}, the client can infer the group ID using gradient-based optimization to find a linear superposition of learned binary masks which minimizes the output entropy. 
~\citet{wortsman2020supermasks} proposed these solutions to learn multiple tasks without catastrophic forgetting in continual learning. 
In these solutions, each client infers the group ID by choosing the most confident binary mask that produces more stable results. There are two variants of this approach as follows: 

\paragraphb{OneShot inference:} Algorithm~\ref{alg:IDE} line 5-9 in Appendix~\ref{Sec:unkownalgs} shows this approach. At \name training round $t$, the server broadcasts $Q$ leaned binary masks $M_{g,q\in[Q]}^{t}$ to the selected clients. Next, each client assigns a confidence coefficient ($\alpha_q$) to each binary mask. 
Each $\alpha_q$ represents how much the client is confident that $q$th binary mask is its match. 
Then it calculates the output of the model as the weighted superposition of these masks (i.e., $p(\alpha) = f \left(x, \theta^w \bigodot (\sum_{q=1}^{Q} \alpha_q M_{g,q}^{t})\right)$).
The $\alpha$ is initialized in a way that all the masks have equal chance ($\alpha_0[q] = \frac{1}{Q}$ for $q \in [Q]$).
Now, Each client tries to find the perfect $\alpha_q$ that minimizes the entropy of the outputs $H(p(\alpha))$ by applying gradient descent with respect to $\alpha$ just for one round, i.e., $\alpha \gets \alpha - \beta \triangledown_{\alpha} H(p(\alpha))$. 
In this approach, the client chooses the group ID $q$ that changing its confidence level ($\alpha_q$) has the most impact on the entropy of the output of mixed models, i.e., $\arg\max_{q} \left( - \frac{\partial H(p(\alpha)))}{\partial \alpha_q}\right)$.





\paragraphb{Binary Search:} Algorithm~\ref{alg:IDE} line 10-23 in Appendix~\ref{Sec:unkownalgs} shows this approach. In this approach, the client utilizes binary search to find the best group ID by removing half of the candidates at each step until one $\alpha_q$ remains nonzero which indicates the best candidate. 
The client calculates the $p(\alpha)$ for $\alpha$, then it applies the gradient descent with respect to $\alpha$ on entropy of $p(\alpha)$. Next, the client eliminates half of the coefficients where they produce gradients less than the median of the gradients.  


\vspace{-6pt}
\section{Experiments}
In this section, we investigate the utility, fairness, and communication cost of \name in two different settings of known and unknown group IDs. 
We use three benchmark datasets widely used in prior works on federated learning application. We run all the experiments for 5 runs with different seeds, and report the average of them. 
Due to space limitations, we defer a detailed discussion of datasets, model architectures, hyperparameters, and the baselines to Appendix~\ref{sec:hyp}. 

\begin{figure}[hbt!]
    \begin{center}
    \begin{subfigmatrix}{1}
    \centering
      \subfigure[Data sample from each group]{
      \includegraphics[width=0.6\columnwidth]{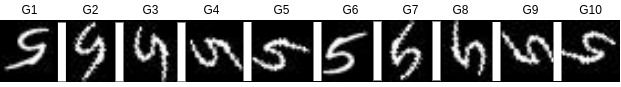}}
    \end{subfigmatrix}    
    \vspace{-6pt}
    \begin{subfigmatrix}{1}
    \centering
      \subfigure[Number of clients in each group]{
      \includegraphics[width=0.9\columnwidth]{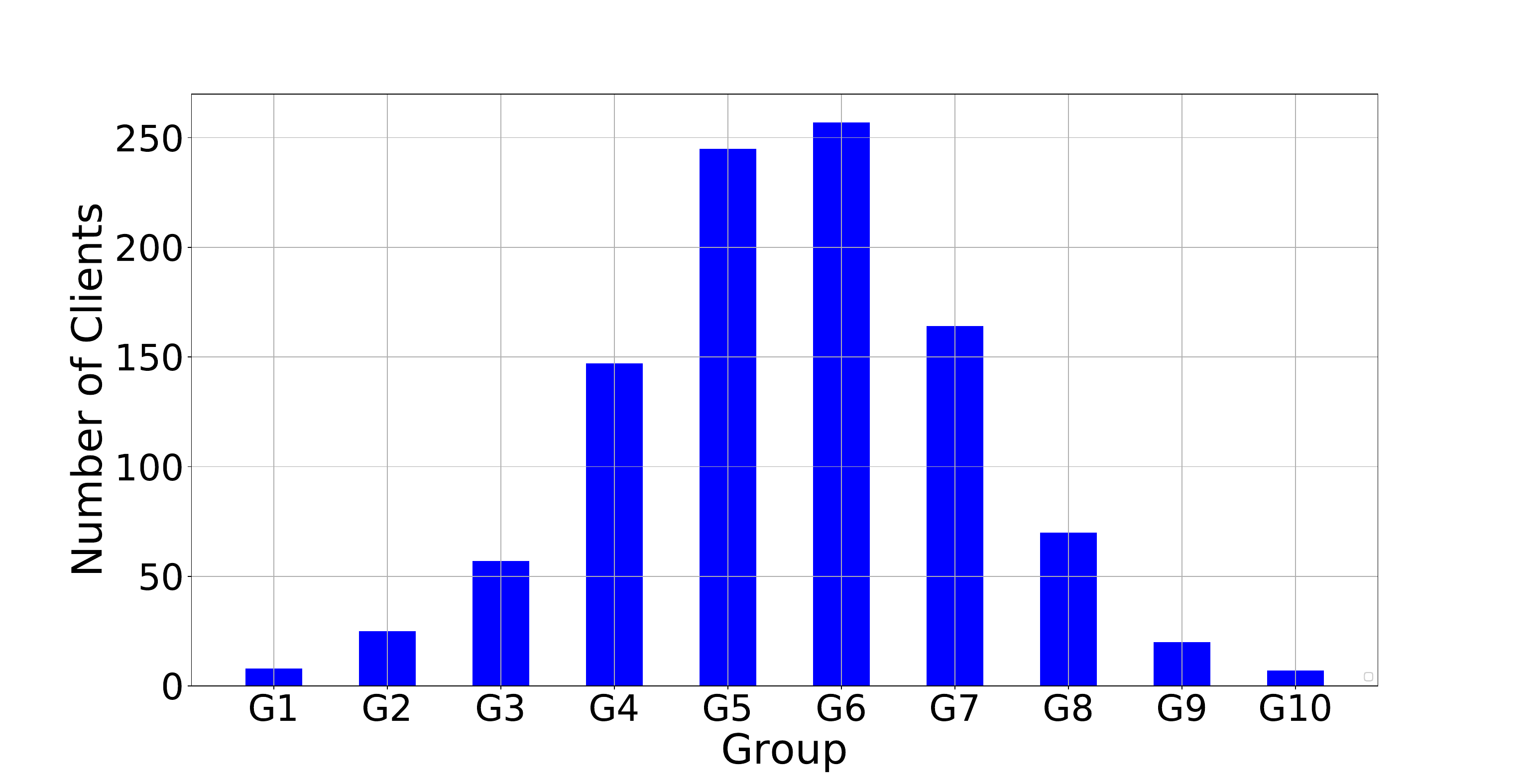}}
    \end{subfigmatrix} 
    \begin{subfigmatrix}{1}
    \centering
      \subfigure[Final test loss for all the groups]{
      \includegraphics[width=0.9\columnwidth]{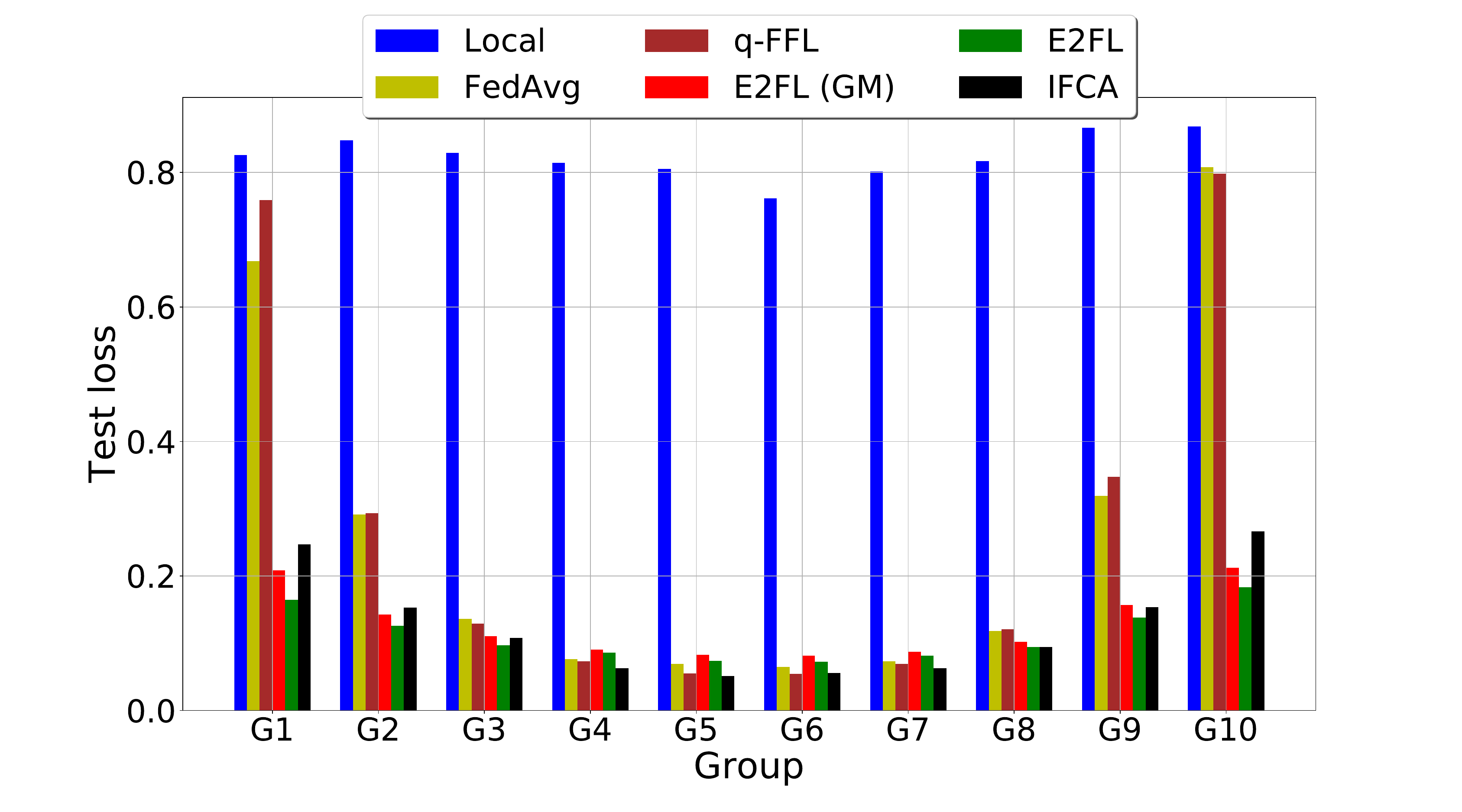}}
    \end{subfigmatrix} 
    \end{center}
    \vspace*{-15pt}
    \caption{FairMNISTRotate: a new dataset to investigate equality and equity in FL application.}
    \label{fig:4Loss1}
\end{figure}

\subsection{Equality vs Equity via \XYZ{}}
\paragraphb{FairMNISTRotate:} To measure the equity and equality, we release a new dataset for fairness experiments in FL application. 
We note that creating different data distributions by manipulating standard datasets such as MNIST has been widely adopted in the continual research community~\citep{goodfellow2013empirical, kirkpatrick2017overcoming, lopez2017gradient}, therefore, we create this dataset by rotating the images in MNIST for each group. Figure~\ref{fig:4Loss1}a shows image samples in each group, and Figure~\ref{fig:4Loss1}b shows the number of clients in each group.
In this dataset, we assign same number of data samples to 1000 clients with 10 different data distributions (with different number of clients in each group). There are majority groups with large number of clients, e.g., G6 with 257 clients, and there are minority groups with small number of clients, e.g., G1 with 8 clients.  In this dataset G1 and G10 are minorities, and G5 and G6 are in majorities.



\begin{table*} [!htb]
\caption{Comparing equality, equity, and communication cost of different variant of FLs on FairMNISTRotate with 1000 clients.} \label{tab:fairMNIST}
\vspace*{-8pt}
\centering
\footnotesize
\begin{tabular} {|c||c|c|c|c||c|c|c|c||c|c|}
  \hline
  \multirow{3}{*}{Approach }  & \multicolumn{10}{c|}{Metric} \\ \cline{2-11}
  
  & \multicolumn{4}{c|}{Group-level Fairness (Equity) }&\multicolumn{4}{c|}{User-level Fairness (Equality) }  & \multicolumn{2}{c|}{Comm Cost} \\ \cline{2-11}
   & Avg  & \vtop{\hbox{\strut Worst}\hbox{\strut (10\%)}} & \vtop{\hbox{\strut Best}\hbox{\strut (10\%)}} & Variance & Avg & \vtop{\hbox{\strut Worst}\hbox{\strut (10\%)}} & \vtop{\hbox{\strut Best}\hbox{\strut (10\%)}} & Variance & \vtop{\hbox{\strut Up}\hbox{\strut (MB)}} & \vtop{\hbox{\strut Down}\hbox{\strut (MB)}} \\ \cline{1-11}
   
  Local training & 84.78  & 84.28 & 85.35  & 0.11 & 85.03  & 81.36  & 87.78 & 3.44 & 0 & 0 \\ \cline{1-11}
     FedAvg &  93.89 & 81.88 & 98.32  & 31.81 & 97.61  & 92.87 & 98.32  & 4.49 & 6.20 & 6.20\\ \cline{1-11}
  IFCA    &  \textbf{97.78}  & \textbf{95.01} & \textbf{99.08}  &  1.98 & \textbf{98.79}  & \textbf{96.86}  & \textbf{99.08}  & 0.35  & 6.20  &  62.0 \\ \cline{1-11}
  q-FFL &  92.23 & 77.31 & 98.42 & 57.46 & 97.56  & 93.33  & 98.42 &  4.32 & 6.20 & 6.20 \\ \cline{1-11}
  \textbf{Our \name} & 96.52  & 93.90  & 97.81 & \textbf{1.63} & 97.48 & 96.07 & 97.81  &  \textbf{0.33} & \textbf{4.05}  & \textbf{5.99} \\ \cline{1-11} 
\hline 
\end{tabular}
\label{tab:MNISTPerm}
\end{table*}

\begin{table*} [!htb]
\caption{Comparing utility, equality, and communication cost on FEMNIST with 3400 default clients.} \label{tab:FEM}
\vspace*{-8pt}
\centering
\footnotesize
\begin{tabular} {|c||c|c|c|c||c|c||}
  \hline
  \multirow{3}{*}{Approach }  & \multicolumn{6}{c|}{Metric} \\ \cline{2-7}
  
  & \multicolumn{4}{c|}{User-level Fairness (Equality) } & \multicolumn{2}{c|}{Communication Cost} \\ \cline{2-7}
   & Average  & \vtop{\hbox{\strut Worst}\hbox{\strut (10\%)}} & \vtop{\hbox{\strut Best}\hbox{\strut (10\%)}} & Variance & \vtop{\hbox{\strut Up}\hbox{\strut (MB)}} & \vtop{\hbox{\strut Down}\hbox{\strut (MB)}} \\ \cline{1-7}
   
  Local training &  68.74 & 44.45  & 87.41 & 154.50 & 0 & 0 \\ \cline{1-7}
     FedAvg & 85.50 & 63.51  & 99.39  & 108.24 & 6.23 & 6.23\\ \cline{1-7}
  q-FFL & 84.40  &64.09  & 99.14  & 100.80 & 6.23 & 6.23 \\ \cline{1-7}
    \textbf{Our \name (OneShot)} & 83.52   & 60.37  & 98.31 & 121.44 & 4.06  & 6.01 \\ \cline{1-7} 
    \textbf{Our \name (Rank Clustering)} &  87.40 & 67.30 & \textbf{100} &   88.54 &  \textbf{4.06 } & \textbf{4.06} \\ \cline{1-7}
  \textbf{Our \name (Lowest Loss)} & \textbf{87.93}  & \textbf{68.59}  & 99.96  & \textbf{81.88}  & 4.06  & 6.01 \\ \cline{1-7}

\hline \hline
\end{tabular}

\end{table*}


In Figure~\ref{fig:4Loss1}c, we compare the performance of different FL algorithms by plotting the test loss of the global trained model for all the ten groups. 
In addition to the results of \name, which is the performance of the global ranking trained for each group, we report the results of the global model as \name (GM).
We additionally compare the utility, equity, equality, and communication cost of these baselines in Table~\ref{tab:fairMNIST}.  

Our experimental results on FairMNISTRotate show that: 
\textbf{(1)}\textbf{ clients have motivation to participate in FL}. All the groups including minorities and majorities get benefit by participating in an FL framework. If each client wants to learn a local model on its local data, they cannot use other clients' knowledge, so their local models perform poorly. 
\textbf{(2)} \textbf{FedAvg gives more attention to majority groups.} FedAvg focuses on the clients from majority groups. Client from majority groups can get more benefit by participating in FedAvg as they have more chance to be selected in each round, so they have more impact on the global model. 
FedAvg can achieve 97.61\% mean test accuracy for all the individual clients (i.e., user-level fairness), but the mean of accuracies for groups is low as 93.89\% which shows that this learning paradigm is focusing on user-level fairness more than on group-based fairness (equity).
\textbf{(3)} \textbf{q-FFL improves equality while worsens equity}. q-FFL is helping the majority groups by ignoring the minorities. q-FFL is a user-level fairness framework, so it makes the results more fair compared to FedAvg in equality; however it produces more unfair results compared in equity. 
\textbf{(4)} \textbf{Training 10 different FLs (i.e., IFCA) is not the best situation for the minorities}. It is important that all the groups in FL share their knowledge.  Figure~\ref{fig:4Loss1}c shows that groups G1, G2, and G10 cannot get similar benefits by participating in IFCA since there is no shared knowledge, and these clients have access to limited data. There is another drawback for IFCA compared to \name which is communication cost. \name clients need to get the binary masks compared to IFCA that they need to get actual weight parameters so \name consumes 5.99 MB compared to IFCA which needs 62.0 MB for download bandwidth. Additionally, IFCA cannot get benefit by our entropy approaches since they are designed for binary masks. 
\textbf{(5)} \textbf{\name is providing equality and equity}. While q-FFL reduces the variance of accuracies for all the clients by 4\% while it increases the variance between groups by 81\% compared to FedAvg. On the other hand our algorithm can reduce both variance of clients and groups by 93\% and 95\% respectively compared to FedAvg

\vspace{-3pt}
\subsection{\name when Group IDs are Unknown}
In this section, we provide experimental results on the FEMNIST~\citep{DBLP:journals/corr/abs-1812-01097} which is a character recognition classification task distributed over 3,400 clients. 
At first, it seems that data distribution among all the clients are similar as all of them are classifying handwritten letters or digits, but there might be hidden groups of clients among these 3400 clients with even more similar handwriting styles.  In this section, we report the performance of \name with different group inference approaches.

In Table~\ref{tab:FEM}, we compare the performance and fairness of different FL algorithms on FEMNIST. We show the user-level fairness (equality) metrics, as there are not specific groups defined for this dataset.
Our experimental results show that:
\textbf{(1)} Training on local data compared to FedAvg does not provide utility, thus all the clients have incentive to  participate in the FL to get more accurate models. \textbf{(2)} q-FFL~\cite{li2019fair} can provide more equal results by reducing the variance by 7\% with cost of reducing the accuracy by 1.10\%. The reason behind this accuracy reduction is that q-FFL gives more attention to making the clients performances more uniform for the clients in majority groups, so it is using smaller portion of the knowledge from minorities . 
\textbf{(3)} \name shows clear advantage over other algorithms, it can provide both utility and fairness. As the groups are unknown in FEMNIST, we use three different approaches of group inference. For Lowest Loss and OneShot approaches, we use 2 static groups, and for rank clustering we use 5 clusters. We also report more details of 2, 3, 4, and 5 clusters in Appendix~\ref{sec:rankclusterExp} Table~\ref{tab:Kmeans2}. 
\name has good features from both worlds: it can improves the utility and fairness together: EFFL increases the average of accuracies by 2.5\% and reduces the variance of clients by 24\%. These benefits are coming from a) \emph{training multiple models each for one group}, and b) \emph{using knowledge of all the participating clients for all the clients by combining all the group models into one global model}. \textbf{(4)} Lowest Loss and rank clustering perform better than OneShot inference approach because in OneShot we have just one forward and backward passes compared to other methods that needs more operations. 

\paragraphb{Miscellaneous Discussions}
Due to space limitations, we defer detailed discussion of ablation studies of \name to Appendix.
In Appendix~\ref{sec:adultexp}, we consider adult census income dataset which is commonly used in fair machine learning literature. In this experiment, we give data samples of different groups to different clients. In Appendix~\ref{sec:rankexpMNIST}, we evaluate the performance of different group inference approaches we introduced on FairMNISTRotate and FEMNIST.

\vspace{-5pt}
\section{Conclusions}
In this paper, we look at the fairness issue in FL with two different lenses. First we define equality and equity as fairness in user-level and group-level, respectively. 
We designed a novel collaborative learning algorithm, called \textbf{E}qual and \textbf{E}quitable \textbf{F}ederated \textbf{L}earning (\name) to achieve both equality and equity.
We validate the efficiency and fairness of \name in different real-world FL applications, and show that \name outperforms existing baselines in terms of the resulting efficiency, fairness of different groups, and fairness among all individual clients.



\section*{Acknowledgements}
This work was supported by NSF grant 2131910.

\bibliography{main}
\appendix
\appendix

\section{Fairness with Multiple Groups in Each Client}~\label{sec:adultexp}
We also provide additional experimental results on a real-world dataset Adult Census Income Dataset~\citep{kohavi1996scaling}.
This dataset contains 48,842 samples extracted from the United States Census Database and classifies whether individuals earn more or less than 50K per year. Prediction values are mapped to 0 $(\leq 50$k$)$ and 1 $(>50$k$)$ where output of $\hat{Y}=1$ is regarded as a positive output (i.e., making more money). 
We consider gender as the protected attribute where we consider male samples $(A=1)$ as the privileged group and female samples $(A=0)$ as unprivileged users. We split the data into train and test, and Table~\ref{tab:Adult} shows the bias in this dataset for difference in their opportunities for making higher income $(Pr[\hat{Y}=1|A=a]$ where $a \in \{0,1\})$. This dataset is biased towards male group where the male group has a higher chance of 31.4\% of getting $\hat{Y}=1$ while female group has a chance of 11.3\% of getting a positive prediction. The models that are trained on these data become more biased towards the male samples in test time by predicting $\hat{Y}=1$ with a higher chance towards the male data input.  

\begin{table}[H]
\caption{Training and test data for male and female samples on Adult dataset.} \label{tab:Adult}
\centering
\footnotesize

\begin{tabular} {|c|c||c|c|}
  \hline
   \multirow{1}{*}{Protected Attr}  & \multirow{1}{*}{Stats} & train data & test data \\ \cline{1-4}
   \multirow{4}{*}{Gender} & Pr$[A=1]$& 67.50\% & 67.50\%  \\ \cline{2-4}
   & Pr$[A=0]$ & 32.5\% & 32.5\% \\ \cline{2-4}
  & Pr$[\hat{Y}=1 | A=1]$ & \textbf{31.4\%} & \textbf{30.8\%} \\ \cline{2-4}
  & Pr$[\hat{Y}=1 | A=0]$ & \textbf{11.3\%} & \textbf{11.3\% }\\ \cline{1-4}
\end{tabular}
\end{table}

We distribute the data samples among 5 clients using Dirichlet distribution with two settings: a) independent and identically distributed (IID) with Dirichlet parameter of $\alpha=5000$, and b) non-independent and identically distributed (non-IID) with Dirichlet parameter of $\alpha=1$.
We use two metrics to measure the fairness in this dataset that are used in previous works~\cite{ezzeldin2021fairfed, abay2020mitigating, zhang2020fairfl}. First, we use equal opportunity difference (EOD) (i.e., $EOD=Pr(\hat{Y}=1|A=0, Y=1)-Pr(\hat{Y}=1|A=1, Y=1)$) where it measures the true positive rate difference of majority and minority group. Second, we use discrimination index (DI) (i.e., $DI=F1(\theta | A=0)- F1(\theta | A=1)$) where it measures the F1 score difference between two groups. 

\begin{table}[hbt!]
\caption{ Fairness of \name compared to other baselines on Adult dataset.} \label{tab:adult1}
\centering
\footnotesize
\begin{tabular} {|c|c||c|c|}
  \hline
   \multirow{2}{*}{Algorithm}& \multirow{2}{*}{Metric} & \multicolumn{2}{c|}{Heterogeneity Level $\alpha$} \\ \cline{3-4}
   & & 5000 (IID) & 1 (Non-IID) \\ \cline{1-4}
  \multirow{3}{*}{FedAvg} & Test Accuracy & 85.56 & 85.47 \\ \cline{2-4}
   & EOD$_{te}$ & -0.0689 & -0.0834\\ \cline{2-4}
   & DI$_{te}$ & -0.0432 & -0.0517 \\ \cline{1-4}
  \multirow{3}{*}{FairFed} & Test Accuracy & 85.1 & 84.47\\ \cline{2-4}
   & EOD$_{te}$ & -0.0701 &-0.069  \\ \cline{2-4}
   & DI$_{te}$ & -0.0441& -0.041 \\ \cline{1-4}
    \multirow{3}{*}{\XYZ{}} & Test Accuracy & 85.22  & 85.20\\ \cline{2-4}
   & EOD$_{te}$ &-0.0174 & -0.0222 \\ \cline{2-4}
   & DI$_{te}$ & -0.019& -0.0252 \\ \cline{1-4} 
\end{tabular}
\end{table}

Table~\ref{tab:adult1} shows the fairness comparison between FedAvg~\cite{konevcny2016federated, mcmahan2017communication}, FairFed~\cite{ezzeldin2021fairfed}, and \name for different data distributions. 
We choose FairFed as a baseline as it provides group fairness when we have different data groups (protected attributes) at each client, similar to what we have in the Adult dataset situation. This algorithm adaptively modifies the aggregation weights at the server in each round. The weights are based on the mismatch between the global fairness measure (at the server) and local fairness measure at each client. This algorithm is favouring clients whose local measures match more with the global fairness measure.
This table shows \name reduces the equal opportunity difference (EOD) and discrimination index (DI) of male and female groups with a small cost on the final test accuracy. 
In particular, for non-IID data distribution, \name improves the EOD by 73\% and DI by 51\%, with the negligible cost of losing test accuracy (0.27\%) compared to FedAvg, while FairFed improves EOD by 17\% and DI by 20\% with reducing the test accuracy by 1\% compared to FedAvg. 
From this table, we can see that FairFed cannot provide the same improvement when data is IID distributed as its design is based on very heterogeneous data distribution, while \name achieves similar improvement in both cases.
These results show that \name enforces the model to act more fair even when each client has a combination of all the groups. 
\section{Missing Experiments when Group IDs are Unknown}\label{sec:exp3}

\paragraphb{Missing experiments for rank clustering \name on FairMNISTRotate:}\label{sec:rankexpMNIST}
Table~\ref{tab:Kmeans1} shows the accuracy of our rank clustering approach for the local rankings in \name. In this table we show the average of 10 different runs. In this approach, each E2FL client learns a local ranking for $E=2$ local epochs on it local data at the beginning of E2FL and send the local ranking to the server, so the server can assign different group IDs to the clients. From this table we can see that with larger rankings, we can predict group ID with higher accuracy.


\begin{table}[hbt!]
\caption{Accuracy of group inference for rank clustering on FairMNISTRotate with 1000 clients by collecting local rankings that are trained for 2 local epochs. We show the accuracy of right prediction by using rankings of different layers with different number of parameters.} \label{tab:Kmeans1}
\centering
\footnotesize
\begin{tabular} {c|c||c}
  \hline
  Layer & No Params  & Accuracy of predicting the right group (\%) \\ \cline{1-3}
     Conv1 & 288 &  91.78  \\ \cline{1-3}
     Conv2 & 18432 &  94.94  \\ \cline{1-3}
     FC1 & 1605632 &  93.76  \\ \cline{1-3}
     FC2 & 1280 &  91.99  \\ \cline{1-3}
     ALL & 1625632 &  93.21 \\ \cline{1-3}
\hline \hline
\end{tabular}
\end{table}

\paragraphb{Missing experiments for rank clustering \name on FEMNIST:}\label{sec:rankclusterExp}
Table~\ref{tab:Kmeans2} shows the equity and equality measurements of using \name on FEMNIST for different numbers of clusters. We can see by increasing the number of clusters, we can cover more diverse groups so the results are more fair.

\begin{table*}[hbt!]
\caption{ Comparing utility, equality and equity of \name using rank clustering for different numbers of clusters on FEMNIST with 3400 default clients.} \label{tab:Kmeans2}
\centering
\footnotesize
\begin{tabular} {|c|c||c|c|c|c||c|c|c|c||}
  \hline
  \multirow{3}{*}{Approach } & \multirow{3}{*}{\vtop{\hbox{\strut Number of}\hbox{\strut Clusters}}} &  \multicolumn{8}{c|}{Metric} \\ \cline{3-10}
  
  & & \multicolumn{4}{c|}{Group-level Fairness (Equity) }&\multicolumn{4}{c|}{User-level Fairness (Equality) } \\ \cline{3-10}
   & & Average  & \vtop{\hbox{\strut Worst}\hbox{\strut group}} & \vtop{\hbox{\strut Best}\hbox{\strut group}} & Variance & Average & \vtop{\hbox{\strut Worst}\hbox{\strut (10\%)}} & \vtop{\hbox{\strut Best}\hbox{\strut (10\%)}} & Variance \\ \cline{1-10}
   
    \multirow{4}{*}{\XYZ{}} & 2 & 88.12 & 85.65 & 90.60 & 6.10 & 87.21 & 66.36 & 99.87 & 94.09 \\ \cline{2-10}
    & 3 & 88.03 & 84.74 & 91.33 & 10.82 & 86.36 & 64.82 & 99.66 & 101.80 \\ \cline{2-10}
    & 4 & 88.01 & 85.01 & 91.39 & 5.42 & 87.12 & 65.81 & 100 & 98.80 \\ \cline{2-10}
    & 5 & 87.61 & 85.31 & 91.06 &  3.80 & 87.40 & 67.30 & 100 &   88.54 \\ \cline{2-10}
\hline \hline
\end{tabular}
\end{table*}

\begin{figure}[hbt!]
    \centering
    \includegraphics[width=0.49\textwidth]{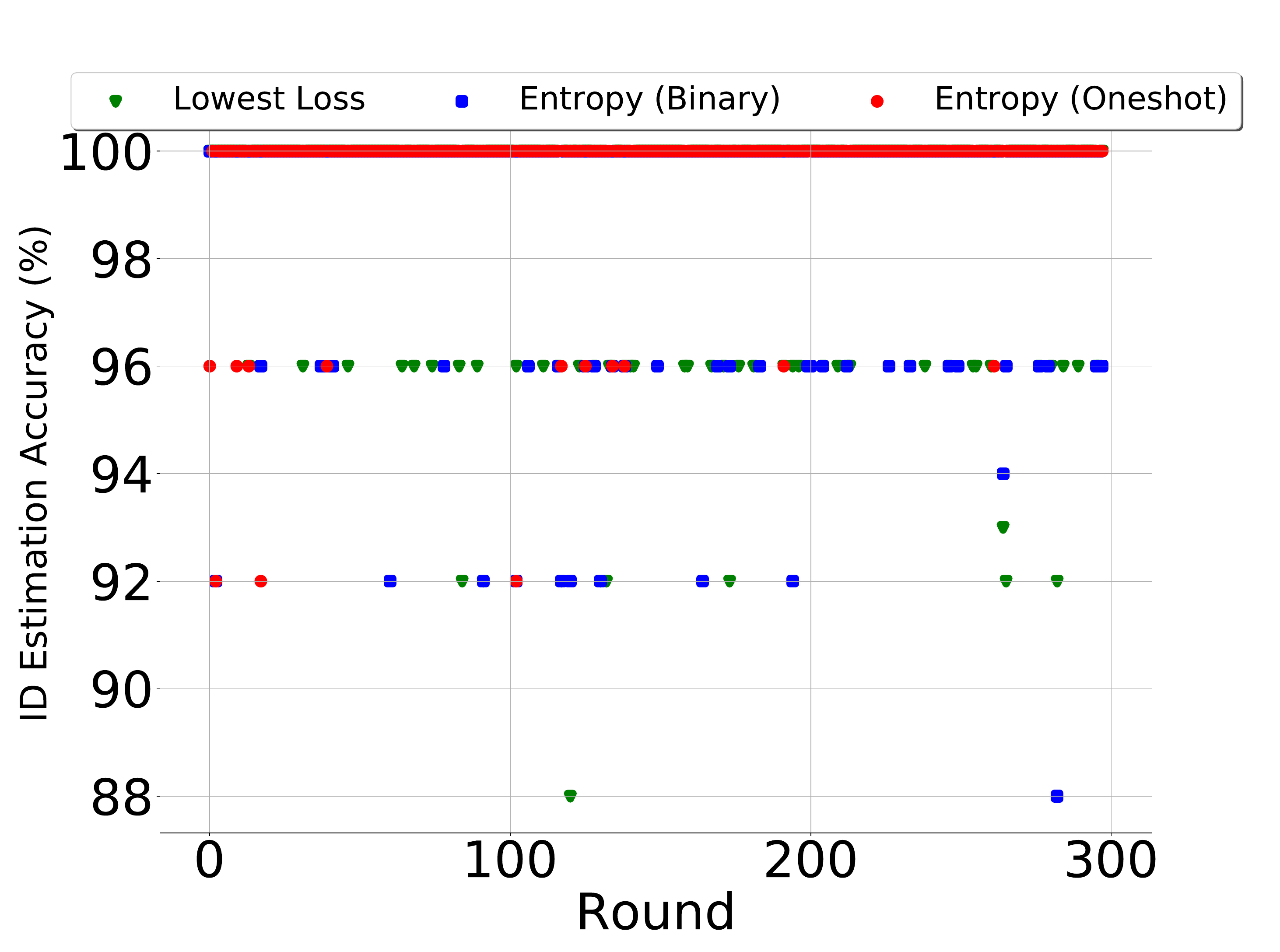}
    \vspace*{-10pt}
    \caption{Accuracy of group inference approaches proposed for \name for 300 initial global epochs on FairMNISTRotate}
    \label{fig:Acc1}
\end{figure}

\paragraphb{Missing experiments for client-based group inference approaches:}\label{sec:client_based_exp}
Figure~\ref{fig:Acc1} shows the accuracy of group inference approaches by using three methods we proposed (on the client-side) for the 300 initial global epochs of \name on the FairMNISTRotate. We can see that the Lowest Loss approach produces more accurate results compared to Binary and OneShot approaches with cost of calculating $Q$ forward passes. Also, the Binary method can produce better results compared to OneShot as it requires more forward and backward passes to determine the group ID. 

\section{Missing Details of Group Inference Algorithms} \label{Sec:unkownalgs}
In this section, we discuss the approaches proposed in Section~\ref{Sec:main_unkown} for E2FL group inference.
Algorithm~\ref{alg:IDEKmeans} shows how rank clustering at server works. 
Algorithm~\ref{alg:IDE} shows three approaches we suggest to be used at client-side for group inference.

\paragraphb{Time complexity comparison:}
Lowest loss approach needs to have $\mathcal{O}(Q)$ forward passes to find the binary mask that produces the lowest loss for $Q$ groups. Binary search over the entropy needs to have $\mathcal{O}(\log(Q))$ forward and backward passes. Finally, the OneShot inference only requires $\mathcal{O}(1)$ forward and backward passes, so it makes the estimation process very fast.

\begin{algorithm}[hbt!]
    \footnotesize
    \caption{Identity Inference with Rank Clustering}\label{alg:IDEKmeans}
    \begin{algorithmic}[1]
        \State \textbf{Input:} number of clients $N$, Local rankings $R_{\{i \in [N]\}}$, number of clusters $Q$, number of iterations $T$

        \Function{\mtasks{RankClustering}}{$R_{\{i \in [N]\}}$}
        \State $\textsc{Centroids} \gets$ pick $Q$ random rankings from  $R_{\{i \in [N]\}}$
        \State $r \gets 0$ \Comment{iteration counter}
        \While{$r<T$} 
            \State $\textsc{ClustersRanking} \gets[[] \; \text{for} \; q \; \in [Q])]$
        \For{$u \in [N]$}
            \State $q \gets \textsc{getClosestCluster}(\textsc{Centroids}, R_i)$
            \State $\textsc{ClustersRanking}[q].\text{append}(u)$
        \EndFor
        \State $\textsc{Cnetroids} \gets \textsc{FRLVote}(\textsc{ClustersRanking})$
        \State $r+=1$
        \EndWhile
        \State \textbf{return} $\textsc{ClustersRanking}$
        \EndFunction

    \end{algorithmic}
\end{algorithm}

        \begin{algorithm}[hbt!]
            \footnotesize
            \caption{Identity Inference at Client Side}\label{alg:IDE}
            \begin{algorithmic}[1]
                \State \textbf{Input:} training data $D_u^{tr}$, random weights $\theta^w$, number of groups $Q$, group binary masks $M_{g, q \in [Q]}^t$, loss function $loss(.)$

                \Function{\mtasks{lowestLoss}}{$\theta^w, Q, M_{g, q \in [Q]}^t, D_u^{tr}$}
                \State \textbf{return} $\argmin_{q\in[Q]} loss\left(D_u^{tr}, \theta^w \bigodot M_{g, q \in [Q]}^t\right)$
                \EndFunction
                
                \Function{\mtasks{OneShot}}{$\theta^w, Q, M_{g, q \in [Q]}^t, D_u^{tr}$}
                \State $\alpha \gets [\frac{1}{Q}, \frac{1}{Q}, ... , \frac{1}{Q}]$
                \State $p(\alpha) \gets f \left(D_u^{tr}, \theta^w \bigodot (\sum_{q=1}^{Q} \alpha_q M_{g,q}^t)\right)$
                \State \textbf{return} $\argmax_{q\in[Q]} \left( - \frac{\partial H(p(\alpha)))}{\partial \alpha_q}\right)$
                \EndFunction

                \Function{\mtasks{Binary}}{$\theta^w, Q, M_{g, q \in [Q]}^t, D_u^{tr}$}
                    \State $\alpha \gets [\frac{1}{Q}, \frac{1}{Q}, ... , \frac{1}{Q}]$
                    \While{$||\alpha||_0>1$}
                        \State $p \gets f \left(D_u^{tr}, \theta^w \bigodot (\sum_{q=1}^{Q} \alpha_q M_{g,q}^t)\right)$
                        \State $g \gets \triangledown_{\alpha} H(p)$
                        \For{$q \in [Q]$}
                            \If{$g_q\leq$ median($g$)}
                                \State $\alpha_q \gets 0$
                            \EndIf
                        \EndFor
                        \State $\alpha \gets \alpha/||\alpha||_1$
                    \EndWhile
                
                \State \textbf{return} $\argmax_{q\in[Q]} \left( \alpha_q \right)$
                \EndFunction

            \end{algorithmic}
        \end{algorithm}
\section{Missing Details of Federated Rank Learning (FRL)}\label{sec:FRL}
\paragraphb{Supermask Learning:}
Modern neural networks have a very large number of parameters. These networks are generally overparameterized~\cite{dauphin2013big, denil2013predicting, LotteryFL, li2021fedmask}, i.e., they have more parameters than they need to perform a particular task, e.g., classification.
The \emph{lottery ticket hypothesis}~\cite{frankle2018lottery} states that a  fully-trained neural network, i.e., \emph{supernetwork}, contains sparse \emph{subnetworks}, i.e., subsets of all neurons in the supernetwork, which can be trained from scratch (i.e., by training same initialized weights of the subnetwork) and achieve performances close to the fully trained supernetwork.
The lottery ticket hypothesis allows for massive reductions in the sizes of neural networks. Ramanujan et al.~\cite{ramanujan2020what} offer a complementary conjecture that an overparameterized neural network with randomly initialized weights contains subnetworks which perform as good as the fully trained network.
To do so, we should identify a supermask $M_g$, which is a binary mask of 1's and 0's, that is superimposed on the random neural network to obtain the final subnetwork, i.e., $\theta^w \bigodot M_g$ where $\theta^w$ is showing the weight parameters for supernetwork.

\paragraphb{Federated Rank Learning (FRL):}
In this section, we provide the design of federated rank learning (FRL) algorithm~\cite{mozaffari2021frl}. 
FRL clients collaborate (without sharing their local data) to \emph{find a subnetwork} within a randomly initialized, untrained neural network called the \emph{supernetwork}.
Algorithm~\ref{alg:FRL} describes  FRL's training. Training a global model in FRL means first finding a  unanimous ranking of supernetwork edges and then using the subnetwork of the top-ranked edges as the final output.

\begin{algorithm}[hbt!]
\caption{Federated Rank Learning (FRL)}\label{alg:FRL}
\begin{algorithmic}[1]
\State \textbf{Input:} number of FL rounds $T$, number of local epochs $E$, number of selected users in each round $n$, seed \textsc{seed}, learning rate $\eta$, subnetwork size $k$\%

\State \serverbox{Server: Initialization}
\State $\theta^s, \theta^w \gets$ Initialize random scores and weights of global model $\theta$ using \textsc{seed}
\State $R_{g}^{1} \gets \textsc{ArgSort}(\theta^s)$ \Comment{Sort the initial scores and obtain initial rankings}

\For{$t \in [1,T]$}
    \State $U \gets$ set of $n$ randomly selected clients out of $N$ total clients
    \For{$u$ in $U$}
        \State \clientbox{Clients: Calculating the ranks}
        \State $\theta^s, \theta^w \gets $ Initialize scores and weights using \textsc{seed}
        \State $\theta^s[R_{g}^{t}] \gets \textsc{sort}(\theta^s)$ \Comment{sort the scores based on the global ranking}
        \State $S \gets$ Edge-PopUp($E, D_u^{tr}, \theta^w, \theta^s, k, \eta$) \Comment{Client u uses Algorithm\ref{alg:edgepop} to train a supermask on its local training data}
        \State $R_{u}^{t} \gets \textsc{ArgSort}(S)$ \Comment{Ranking of the client}
    \EndFor
    \State \serverbox{Server: Majority Vote}
    \State $R_{g}^{t+1} \gets \textsc{Vote}(R_{u \in U}^{t})$ \Comment{Majority vote aggregation}
\EndFor

\Function{Vote}{$R_{\{u \in U\}}$ }:
    \State $V \gets \textsc{ArgSort}(R_{\{u \in U\}})$
    \State $A \gets \textsc{Sum}(V)$
    \State \textbf{return} $\textsc{ArgSort}(A)$
\EndFunction
\end{algorithmic}
\end{algorithm}

\paragraphb{Edge-popup Algorithm}~\label{sec:EP}
 Algorithm~\ref{alg:edgepop} presents Edge-popup algorithm~\cite{ramanujan2020what}.

        \begin{algorithm}[hbt!] 
            \caption{Edge-popup (EP) algorithm: it finds a subnetwork of size $k$\% of the entire network $\theta$} \label{alg:edgepop}
            \begin{algorithmic}[1] 
            \State \textbf{Input:} number of local epochs $E$, training data $D$, initial weights $\theta^w$ and scores $\theta^s$, subnetwork size $k$\%, learning rate $\eta$
            \For{$e \in [E]$}
            \State $\mathcal{B}\gets$ Split $D$ in $B$ batches
            \For{batch $b\in[B]$}
                \State \textsc{EP Forward} ($\theta^w, \theta^s, k, b$)
                \State $\theta^s=\theta^s - \eta\nabla \ell(\theta^s;b)$
            \EndFor
        \EndFor
        \State \textbf{return} $\theta^s$
                \Function{EP forward}{$\theta^w, \theta^s, k, b$}
                \State $m \gets$ sort$(\theta^s)$
                \State $t \gets$ $int((1-k)* len(m))$
                \State $m[:t]=0$
                \State $m[t:]=1$
                \State $\theta^p =\theta^w \odot \mathbf{m}$
                \State \textbf{return} $\theta^p(b)$
                \EndFunction
                
            \end{algorithmic}
        \end{algorithm}
\section{Missing Details of Experiment Setup}\label{sec:hyp}

\subsection{Datasets and Model Architectures}
In this section, we discuss our hyperparameters for each baseline and each dataset. 
In all the experiments we use SGD as the optimizer, and we use a momentum of 0.9 with weight decay of 1e-4. 
We tune the learning rate and local epochs for each dataset and each baseline to find the best results. 
We run all the experiments for 3000 global epochs and report the final results. q-FFL converges slower, so we allow 6000 global epochs for q-FFL. 

\begin{table*}[hbt!]
\caption{Model architectures. } \label{tab:models}
\centering
\footnotesize
\begin{tabular} {|c|c|c|}
  \hline
  Architecture & Layer Name & Number of params \\ \hline \hline
 
 \multirow{5}{*}{LeNet (FairMNISTRotate)} & Convolution(32) + Relu &  288 \\ \cline{2-3}
  & Convolution(64) + Relu &  18432 \\ \cline{2-3}
  & MaxPool(2x2) &  - \\ \cline{2-3}
 & FC(128) + Relu &  1605632 \\ \cline{2-3}
  & FC(10) &  1280 \\ \cline{2-3}
\hline \hline 

  \multirow{5}{*}{LeNet (FEMNIST) } & Convolution(32) + Relu &  288 \\ \cline{2-3}
  & Convolution(64) + Relu &  18432 \\ \cline{2-3}
  & MaxPool(2x2) &  - \\ \cline{2-3}
 & FC(128) + Relu &  1605632 \\ \cline{2-3}
  & FC(62) &  7936 \\
  \hline \hline 
    \multirow{3}{*}{FC (Adult)} & FC(1024) + Relu &  103424 \\ \cline{2-3}
  & FC(1024) + Relu &  1048576 \\ \cline{2-3}
 & FC(2)  &  2048 \\ \cline{2-3}
\hline \hline
\end{tabular}
\end{table*}

\paragraphb{FairMNISTRotate:}
We experiment with LeNet architecture given in Table~\ref{tab:models}. For local training in each \name round, each client uses 2 epochs with learning rate of 0.1. For FedAvg, we use 2 local epochs with learning rate of 0.01. For q-FFL, we set the q-FFL fairness hyperparameter to 0.1 along with the same settings of FedAvg. We use batch size of 8, and we select 25 client (out of 1000) in each FL round for all the experiments.

\paragraphb{FEMNIST~\cite{DBLP:journals/corr/abs-1812-01097, DBLP:conf/ijcnn/CohenATS17}}
is a character recognition classification task with 3,400 clients, 62 classes (52 for upper and lower case letters and 10 for digits), and 671,585 grey-scale images. Each client has data of their own handwritten digits or letters. We use LeNet architecture given in Table~\ref{tab:models}. 
For local training in each \name round, each client uses 2 epochs with learning rate of 0.05.
For FedAvg, we use 2 local epochs with learning rate of 0.05. For q-FFL, we set the q-FFL fairness hyperparameter to 0.1 with the same setting of FedAvg.
We use batch size of 10, and we select 25 clients (out of 3400) in each FL round for all the experiments.

\paragraphb{Adult}
Adult Census Income Dataset~\cite{kohavi1996scaling} is a dataset commonly used for fair machine learning literature. 
The dataset consists of anonymous information such as gender, race, education, occupation, etc. We experiment with a simple 3-layer fully connected network given in Table~\ref{tab:models}.
We consider gender as the protected attribute where we consider male samples $(A=1)$ as the privileged group and female samples $(A=0)$ as unprivileged users. 
For experiments on the Adult dataset, we use all of the updates $(n=N=5)$ in each round. We train for 20 global epochs, and in each global epoch, we ask the clients to run for 5 local epochs using SGD as the optimizer with learning rate of 2.0, momentum of 0.9 and weight decay of 1e-4 for all the experiments.


\subsection{Missing Details of Baselines }

\paragraphb{Federated averaging (FedAvg)~\cite{konevcny2016federated, mcmahan2017communication}:} 
FedAvg is an effective aggregation rule (AGR) where due to its efficiency, Average is the only AGR implemented by FL applications in practice~\cite{DBLP:journals/corr/abs-2007-10987, DBLP:journals/corr/abs-2102-08503}. 

\paragraphb{IFCA~\citep{ghosh2020efficient}} 
This algorithm assumes that users are partitioned into different clusters, and their goal is to train a separate model for each cluster. Their main challenge was how to identify the cluster membership of each user. For their experiments, they assume that they know the number of clusters. Also in case of ambiguous cluster structure, they treat the number of clusters as a hyperparameter to tune. They propose Iterative Federated Clustering Algorithm (IFCA) that solves simultaneously two problems: 1) identifies the cluster membership of each user and 2) optimizes each cluster model. In IFCA, the server will send all the models to selected clients in each round, and each client will pick the best model that produces the least amount of loss and participates in the training of that cluster. On the server side, it is just FedAvg for several clusters. 

\paragraphb{Local training:} A framework in which the clients train standalone models on local datasets without collaboration.

\paragraphb{FedFair~\cite{ezzeldin2021fairfed}:} This algorithm adaptively modifies the aggregation weights at the server in each round. The weights are based on the mismatch between the global fairness measure (at the server) and the local fairness measure at each client. This algorithm is favouring clients whose local measures match more with the global fairness measure.

\paragraphb{q-FFL~\cite{li2019fair}:} Inspired by fair resource allocation algorithms for wireless networks, Li et al.~\cite{li2019fair}  design a federated optimization technique called q-FFL. 
This technique aims at improving FL fairness by minimizing the aggregate reweighed loss, in a way that the devices with higher loss are given higher relative weights.

\end{document}